\documentclass[twoside,11pt]{article}

\usepackage{jmlr2e}
\usepackage{subcaption}
\usepackage{multirow}
\usepackage{wrapfig}

\jmlrheading{1}{}{}{}{}{}

\firstpageno{1}

\begin{document}

\title{Random Actions vs Random Policies: Bootstrapping Model-Based Direct Policy Search}

\author{\name Elias Hanna \email h.elias@hotmail.fr \\
       \addr Sorbonne Université, CNRS, \\Institut des Systèmes Intelligents et de Robotique, ISIR, \\F-75005 Paris, France
       \AND
       \name Alex Coninx \email coninx@isir.upmc.fr \\
       \addr Sorbonne Université, CNRS, \\Institut des Systèmes Intelligents et de Robotique, ISIR, \\F-75005 Paris, France
       \AND
       \name Stéphane Doncieux \email doncieux@isir.upmc.fr \\
       \addr Sorbonne Université, CNRS, \\Institut des Systèmes Intelligents et de Robotique, ISIR, \\F-75005 Paris, France
       }

\editor{}

\maketitle

\begin{abstract}%   <- trailing '%' for backward compatibility of .sty file
\textit{This paper studies the impact of the initial data gathering method on the subsequent learning of a dynamics model. Dynamics models approximate the true transition function of a given task, in order to perform policy search directly on the model rather than on the costly real system. This study aims to determine how to bootstrap a model as efficiently as possible, by comparing initialization methods employed in two different policy search frameworks in the literature. The study focuses on the model performance under the episode-based framework of Evolutionary methods using probabilistic ensembles. Experimental results show that various task-dependant factors can be detrimental to each method, suggesting to explore hybrid approaches.}
\end{abstract}


%%%%%%%%%%%%%%%%%%%%%%%%%%%%%%%%%%%%%%%%%%%%%%%%%%%%%%%%%%%%%%%%%%%%%%%%%%%%%%%%%%%%%%
%%%%%%%%%%%%%%%%%%%%%%%%%%%%%%%%%%%%%%%%%%%%%%%%%%%%%%%%%%%%%%%%%%%%%%%%%%%%%%%%%%%%%%
%%%%%%%%%%%%%%%%%%%%%%%%%%%%%%%%%% INTRODUCTION %%%%%%%%%%%%%%%%%%%%%%%%%%%%%%%%%%%%%%
%%%%%%%%%%%%%%%%%%%%%%%%%%%%%%%%%%%%%%%%%%%%%%%%%%%%%%%%%%%%%%%%%%%%%%%%%%%%%%%%%%%%%%
%%%%%%%%%%%%%%%%%%%%%%%%%%%%%%%%%%%%%%%%%%%%%%%%%%%%%%%%%%%%%%%%%%%%%%%%%%%%%%%%%%%%%%

\section{Introduction}

\par
Sparse reward tasks are frequent in robotics and call for data-greedy learning algorithms with strong exploration capabilities \citep{lehman2011abandoning,pugh2016quality,cully2017quality,kim2017learning}. They have recently shown promising results on notoriously difficult tasks like grasping as they have discovered grasping movements without the need of demonstrations or gripper specific constraints \citep{morel2022automatic}. Such learning algorithms are based on direct policy search \citep{sigaud2019policy} and cannot be used directly on the real robot as they need up to millions of policy evaluations. They thus heavily rely on simulations, but simulations are not perfect representations of the real robotic setup. It may create issues when such discrepancies are exploited by the learning algorithm, leading to issues either called reality gap \citep{jakobi1995noise} or simulation bias \citep{an1988model, atkeson1997robot}.

\par
To alleviate this problem, several techniques do exist. Some ignore the badly modeled parts of the simulator, potentially missing the most rewarding behaviours \citep{koos2012transferability}. Sim-to-real approaches, like domain randomization \citep{tobin2017domain} or domain adaptation \citep{jiang2008literature}, are only efficient when the source domain (usually a simulator) dynamics are not too different from the real system dynamics. 
%Such methods thus do not work on systems whose dynamics are not well simulated, \textit{e.g.} deformable objects \citep{billard2019trends}. 
Model-based approaches do not suffer from those issues \citep{polydoros2017survey}, as they directly learn a model of the system from data gathered on the target domain. The robot behaviour is then trained using the model in different fashions.
%, either fully on the learnt model, or using both data gathered on the model and on the real system.

\par
Policy search consists in defining a parameterized policy and in directly optimizing its parameters while it is controlling a robot \citep{sigaud2019policy}. Model-based policy search is a long standing research field made of very diverse methods ranging from methods inspired by model predictive control (MPC) \citep{camacho2013model} to model-based reinforcement learning (RL) \citep{polydoros2017survey}. On one side there is step-based policy search methods, mainly resulting from the reinforcement learning framework \citep{sutton2018reinforcement}, and on the other side, there is episode-based policy search methods, mainly resulting from Bayesian Optimization \citep{snoek2012practical} and from Evolutionary methods \citep{stanley2019designing}. The interest of using a model in the RL framework is not new, and first promising results were obtained in PILCO by \cite{deisenroth2011pilco}. Model-based techniques started by using Gaussian Processes \citep{deisenroth2011pilco,gaier2018data}, but as problems arose \citep{huang2015scalable}, more methods started turning to Neural Networks architectures for dynamics modeling \citep{gal2016improving,nagabandi2018neural,chua2018deep,sharma2019dynamics,lim2021dynamics}.

\par
Model-based policy search is thus promising \citep{gaier2018data,keller2020model,lim2021dynamics,lim2022learning}, but it requires to learn a model that is accurate enough to predict the behavior of the policy over a complete rollout, \textit{i.e}. with model predictions that are used in closed loop over a given horizon. In this case, error accumulates and makes predictions rapidly diverge from the ground truth, thus misleading any learning algorithm that would rely on it. The bootstrap phase in which data is acquired to train a first model is thus critical. As experiments on the real robot are to be minimized, it is important to define the most efficient approach to randomly gather initial data. Two different approaches have been used, either random actions \citep{nagabandi2018neural, hafner2019learning, sekar2020planning} or random policies \citep{lim2021dynamics,lim2022learning}. We compare their impact on the generated data distribution and on the obtained model quality, so that an active model-based learning approach is bootstrapped as efficiently as possible.

\section{Method}

\par
We place ourselves in the Reinforcement Learning framework \citep{sutton2018reinforcement}. In this context, we define the tuple $< \mathcal{S}, \mathcal{A}, f, \pi >$, where $\mathcal{S}$ is the state-space in which the agent can take states $s$ and that contains all the needed information to determine the agent and environment dynamics, $\mathcal{A}$ is the action-space in which the action can draw actions from. $f$ is the transition function that describes how the agent actions influence the state of the system such that $f: \mathcal{A} \times \mathcal{S} \rightarrow \mathcal{S}$, and finally $\pi$ is a function that maps an action to a specific state $s \in \mathcal{S}$ such that $\pi: \mathcal{S} \rightarrow \mathcal{A}$. %Methods that search for a policy to solve a task are thus called policy search methods, and we will particularly be interested in policy search with a model.

\par 
The transition function $f$ is supposed to be deterministic: $s_{t+1} = f(s_t, a_t)$. Learning the dynamics model is the same as learning explicitly the system's transition function $f$, by approximating it with a function $\hat{f}_{\theta}$ parameterized by a vector $\theta$. The idea behind model based policy search is thus to fit a model $\hat{f}_{\theta}$ given limited measurements of the true transition function $f$ in the form of $N$ data samples $\mathcal{D} = \left\{ (s_n, a_n), s_{n+1}\right\}^N_{n=1}$. The approximated function $\hat{f}$ is then used recursively to predict a policy behaviour without interacting with the costly real system. Policies are then rolled out on the model on a given horizon $H$. We call episode a complete policy rollout on the learnt model. In this paper, the model that learns the dynamics of the task is represented by ensembles of probabilistic models \citep{chua2018deep,hafner2019learning, sekar2020planning}. We refer the reader to \cite{chua2018deep} for more details on ensembling for dynamics modeling.

\WFclear
\begin{wrapfigure}{r}[0mm]{.4\textwidth}
    \vspace*{\fill}
    \centering
    {\includegraphics[width=.25\textwidth]{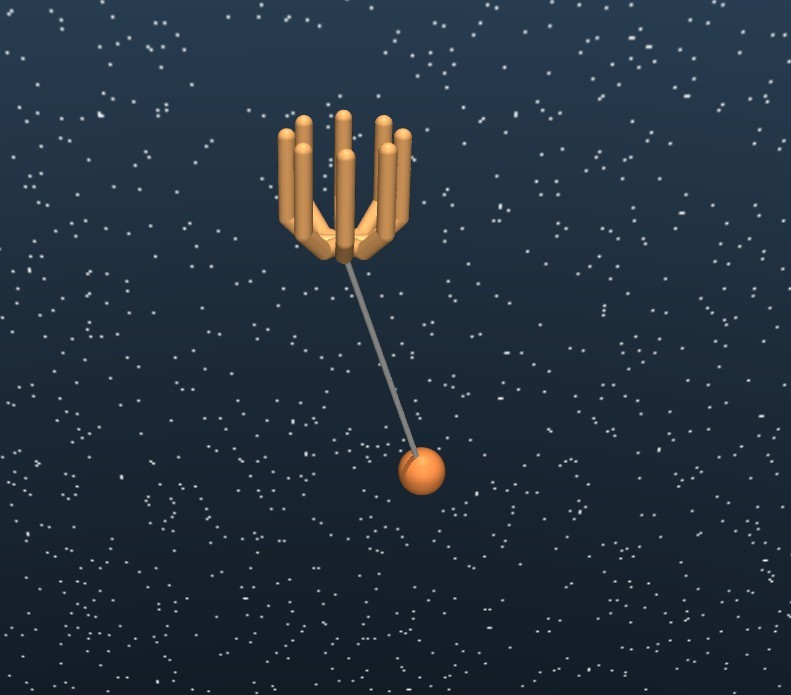}
    \subcaption{Ball In Cup Environment}
    \label{setup_exp:bic}}\par\vfill
    {\includegraphics[width=.25\textwidth]{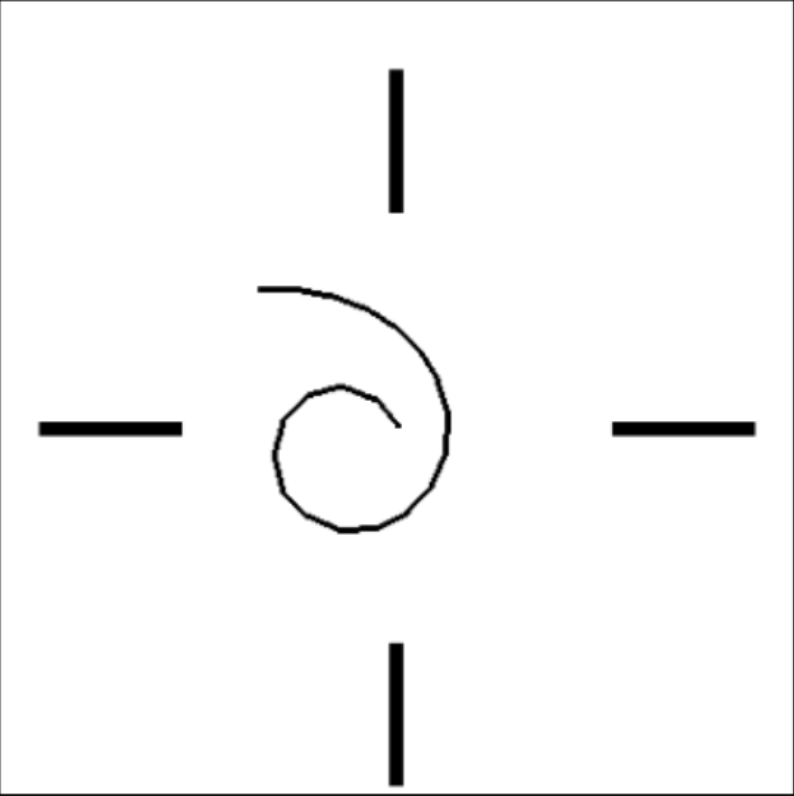}
    \subcaption{Redundant Arm with walls environment}
    \label{setup_exp:ra}}
\label{hist_repart:bic_ac_0}
\captionsetup{justification=centering}
\caption{Benchmark environments}
\vspace{-5mm}
\end{wrapfigure}

\par
As we want to study the influence of the model initial data gathering when coupled with evolutionary methods, we further detail the variables of interest we will be looking into for our experiments. Indeed, evolutionary methods, like Novelty Search \citep{lehman2011abandoning} or Quality-Diversity approaches \citep{lehman2011evolving,pugh2016quality}, make use of what is called a \textit{behavior space}, denoted $\mathcal{B}$, and of an \textit{observer function} $o_{\mathcal{B}}$ which associates a behavior descriptor $b \in \mathcal{B}$ to a trajectory of states $\tau = \left\{s_0, s_1, ..., s_T\right\}$ of length T such that $o_{\mathcal{B}}: S^T \rightarrow \mathcal{B}$. This behavior descriptor $b$ thus characterizes the agent's behavior in a way that is task-aligned and more compact than the whole trajectory $\tau$.
When using dynamics models, the part of most interest for evolutionary methods is thus the ability for the dynamics model to predict the agent's behavior $b$, as the agent's behavior is directly derived from the agent trajectory in the state-space during a rollout.
%SD: il faudrait dire là pourquoi on n'apprends pas directement un modèle qui prédit ce descripteur...

%\par
%We will thus be comparing two initialization methods for model training data gathering that are evaluated in an episodic framework. The first method considered is the selection of an action $a$ by uniformly sampling in the action-space $\mathcal{A}$, which is within the $[-1, 1]$ range in our environments, then repeating that action $R$ times. Actions are sampled until the number of total steps on environment reaches a complete episode horizon $H$. We hypothesize that the obtained policies will cover a narrow portion of the state-space $\mathcal{S}$, but in a quite uniform and dense way. On the other hand, we consider the selection of an action $a$ from a randomly parameterized policy $\pi_{\theta}:\mathcal{S} \rightarrow \mathcal{A}$, represented by a feed forward fully connected neural network with two hidden layers of 10 neural networks each, with relu activations in the hidden units and an hyperbolic tangent activation on the output layer. Policy parameters are uniformly drawn within the $[-5, 5]$ range, and the policy is then used to select an action given the current state of the agent.

\par 
Model-based approaches are iterative and alternate between exploration on the real system and exploration in the model. The focus here is on the initial data gathering that bootstraps the process. Two initialization methods will be compared:
\begin{itemize}
    \item Random policies: randomly parameterized policies, represented as a fully connected neural network with two hidden layers of size 10, rolled out on the task horizon $H$ and taking as input the system current state $s_t$. An episode consists of rolling out such a policy on a newly reinitialized environment.
    \item Random actions: $B$ uniformly drawn actions $a \in \mathcal{A}$, each applied for $R$ step sequentially on the task horizon $H$ such that $B=\frac{H}{R}$. An episode thus consists of rolling out $B$ different actions, for $R$ step each, on a newly reinitialized environment.
\end{itemize}

We also consider an hybrid approach that splits its budget evenly between the two aforementioned methods, which will be called Random Action Random Policies Hybrid (RARPH).

The only parameter we can influence to collect data is the way the agent acts in its environment. Ideally, we would need independent and identically distributed data to train the dynamics model. Indeed, this is a requirement so that the model is able to generalize well once it is used on unseen parts of the real-system state-action space. The issue is that \textit{i.i.d.} actions do not guarantee \textit{i.i.d.} states as the mapping between the two is a complex function. Random policies have been used to generate initial training data with convincing results although inducing a strong bias as all $H$ samples collected during an episode come from a single trajectory and are thus not independent. In contrast, random actions should provide data closer to \textit{i.i.d.} data, as the collected data samples are only dependant by subsets of size $R$ and should be uniformly distributed in the whole action space $\mathcal{A}$. 

The goal of this study is thus to determine if such an induced bias is detrimental or profitable to learning a dynamics model of the task, and to what extent it is task-dependant. We hypothesize that random policies might be advantageous in environments with sparse interactions, as they tend to explore a broader part of the state-space, while random actions should provide a safe initialization method, whatever the environment is. The naive hybrid approach proposed is expected to bring better performance than random actions in environments with sparse interactions, while remaining competitive in others as well.

% The hypothesis that we make here is that random policies will cover a broader part of the state-space $\mathcal{S}$ but in a sparser way. In the case of learning a dynamics model for combined use with episode-based evolutionary methods, we make the hypothesis that such random policies are better fitted for this task, as they seem closer in terms of data distribution to the evolutionary methods collected data.

%%%%%%%%%%%%%%%%%%%%%%%%%%%%%%%%%%%%%%%%%%%%%%%%%%%%%%%%%%%%%%%%%%%%%%%%%%%%%%%%%%%%%%
%%%%%%%%%%%%%%%%%%%%%%%%%%%%%%%%%%%%%%%%%%%%%%%%%%%%%%%%%%%%%%%%%%%%%%%%%%%%%%%%%%%%%%
%%%%%%%%%%%%%%%%%%%%%%%%%%%%%%%%%% EXPERIMENTS %%%%%%%%%%%%%%%%%%%%%%%%%%%%%%%%%%%%%%%
%%%%%%%%%%%%%%%%%%%%%%%%%%%%%%%%%%%%%%%%%%%%%%%%%%%%%%%%%%%%%%%%%%%%%%%%%%%%%%%%%%%%%%
%%%%%%%%%%%%%%%%%%%%%%%%%%%%%%%%%%%%%%%%%%%%%%%%%%%%%%%%%%%%%%%%%%%%%%%%%%%%%%%%%%%%%%

\section{Experiments} 

\subsection{Experimental setups}

%\subsubsection{Ball-in-cup}

\WFclear
\begin{wrapfigure}{r}[0mm]{.5\textwidth}
    % \vspace*{\fill}
    \vspace{-23mm}
    \centering
    \includegraphics[scale=0.48]{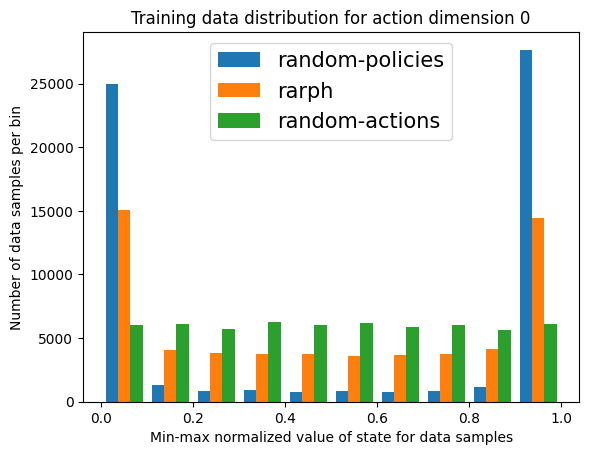}
\captionsetup{justification=centering}
\caption{Histogram of training data distribution on test environments for actions}
\label{hist_repart_ac}
\vspace{-4mm}
\end{wrapfigure}

The first environment considered is the Ball-In-Cup environment (Figure \ref{setup_exp:bic}). It consists of a ball hanging below a cup by a string. The cup is controlled in position in the 3D space. The task consists of putting the ball inside the cup. This problem is interesting as it involves sparse interactions, as giving the ball an upward velocity requires to swing the ball. The state-space consists of the 3D relative position and velocity of the ball to the cup, and the considered outcome space is the relative position of the ball to the cup.

The second environment is the Redundant Arm environment (Figure \ref{setup_exp:ra}). It consists of a 20-DOF robotic arm controlled in an environment with or without obstacles (walls). Each articulation of the robotic arm is torque-controlled. The tasks consist in reaching certain parts of the space with the robot end-effector. The problem is hard to gather data from as hitting a wall or self-colliding stops the episode. Having the two scenarios, with and without walls helps us to distinguish the effect of early stopping on the data gathering of the initialization method. The state-space consists in the position of each of the twenty joints and of the x-y position of the end-effector. The considered outcome space is the end-effector position.

\begin{figure}[h!]
    \begin{subfigure}{.32\textwidth}
        \centering
        \includegraphics[scale=0.3]{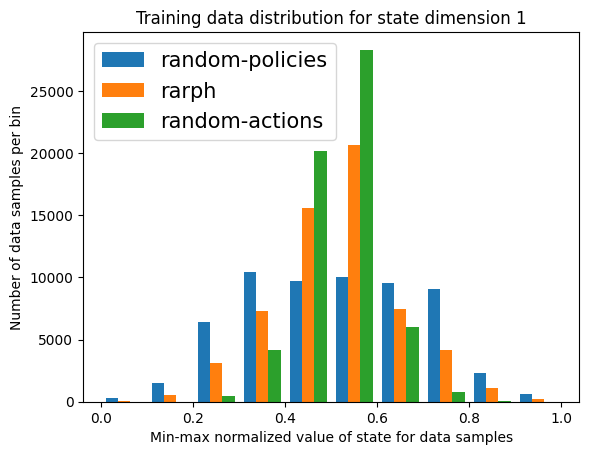}
        \caption{Ball In Cup}
        \label{hist_repart:bic_0}
    \end{subfigure}
    \begin{subfigure}{.32\textwidth}
        \centering
        \includegraphics[scale=0.3]{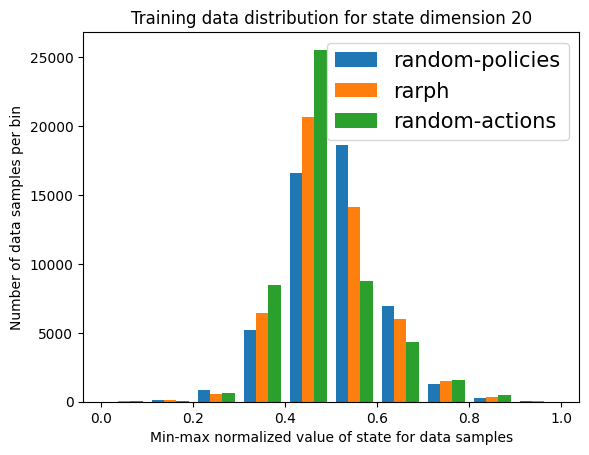}
        \caption{RA No Walls}
        \label{hist_repart:ranw_20}
    \end{subfigure}
    \begin{subfigure}{.32\textwidth}
        \centering
        \includegraphics[scale=0.3]{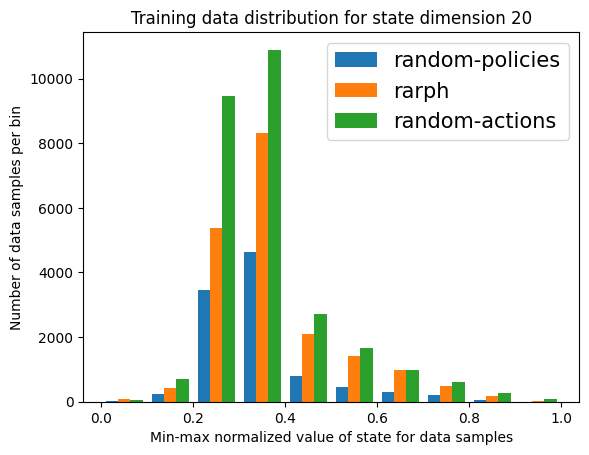}
        \caption{Redundant Arm}
        \label{hist_repart:ra_20}
    \end{subfigure}
    \captionsetup{justification=centering}
    \caption{Histograms of training data distribution on test environments for RARPH}
    \label{hist_repart}
\end{figure}

\subsection{Data distribution analysis}

As stated before, the key interest on the model learning aspect when coupling a learnt model with an evolutionary method is the capacity the model has to predict correctly the agent behavior in its outcome space. Focusing our analysis on the outcome space dimensions of the state-space is thus relevant for this study, as this will highlight which method explores best the state-space dimensions of interest and theoretically lead to less prediction error when using gathered data to train a dynamics model. In order to do so we estimate the discrepancies between the data distributions using histograms of the data distribution on the dimensions of the outcome space for each initialization method.

%%% Ball In Cup %%%

\begin{wraptable}{l}[0mm]{0.7\textwidth}
\vspace{-4mm}

\scalebox{0.6}{
\begin{tabular}{| c | c | c | c | c|}
    \hline
    \multicolumn{5}{| c |}{Ball In Cup}\\\hline
    Initialization episodes & Prediction horizon & Random policies & Random actions & RARPH\\\hline
    \multicolumn{1}{| c |}{\multirow{3}{*}{5}}
    & 1     & 0.022 $\pm$ 0.049 & 0.054 $\pm$ 0.126 & 0.035 $\pm$ 0.061\\\cline{2-5}
    & 20    & 0.349 $\pm$ 0.685 & 0.601 $\pm$ 1.178 & 0.634 $\pm$ 1.052\\\cline{2-5}
    & 300   & 4.140 $\pm$ 2.413 & 26.56 $\pm$ 57.86 & 7.149 $\pm$ 2.591\\\hline
    
    \multicolumn{1}{| c |}{\multirow{3}{*}{10}} 
    & 1     & 0.018 $\pm$ 0.043 & 0.043 $\pm$ 0.093 & 0.030 $\pm$ 0.053\\\cline{2-5}
    & 20    & 0.293 $\pm$ 0.554 & 0.600 $\pm$ 1.175 & 0.521 $\pm$ 0.807\\\cline{2-5}
    & 300   & 3.981 $\pm$ 2.210 & 7.110 $\pm$ 12.36 & 7.003 $\pm$ 2.217\\\hline
    
    \multicolumn{1}{| c |}{\multirow{3}{*}{15}}
    & 1     & 0.018 $\pm$ 0.042 & 0.037 $\pm$ 0.076 & 0.026 $\pm$ 0.050\\\cline{2-5}
    & 20    & 0.284 $\pm$ 0.569 & 0.547 $\pm$ 1.043 & 0.444 $\pm$ 0.755\\\cline{2-5}
    & 300   & 2.880 $\pm$ 1.668 & 4.395 $\pm$ 11.75 & 3.796 $\pm$ 1.603\\\hline
    
    \multicolumn{1}{| c |}{\multirow{3}{*}{20}}
    & 1     & 0.017 $\pm$ 0.042 & 0.033 $\pm$ 0.070 & 0.024 $\pm$ 0.047\\\cline{2-5}
    & 20    & 0.266 $\pm$ 0.537 & 0.520 $\pm$ 1.044 & 0.404 $\pm$ 0.668\\\cline{2-5}
    & 300   & 4.463 $\pm$ 3.080 & 5.467 $\pm$ 4.736 & 4.500 $\pm$ 3.984\\\hline

\end{tabular}}
\captionsetup{justification=centering}
\caption{Mean prediction error on Ball In Cup}
\label{quanti_mean_pred:ball_in_cup}
\vspace{-5mm}
\end{wraptable}

\par
All the results are gathered on each environment with 20 episodes of each method, repeated 10 times. Only one dimension of the outcome space is shown each time, as it is representative of the other outcome space dimensions. As observed on figure \ref{hist_repart:bic_0}, the data distribution of random policies on Ball In Cup is broader, as the agent observes transitions in a wider portion of the outcome space. On contrary, on the Redundant Arm environment we observe on figure \ref{hist_repart:ra_20} that the random actions data distribution is broader and has many more samples due to early stopping. This is explained by the fact that the random policies take action that are more often at the limit of the action space, as shown in figure \ref{hist_repart_ac} (figure obtained on the Ball In Cup environment, but representative of random policies action distribution), which in the case of a torque-controlled robotic arm can lead faster to the joint limits, self-collision or obstacle hitting and thus to episode termination. Removing the walls as shown on Figure \ref{hist_repart:ranw_20} indeed qualitatively brings the two data distributions very close one to another. As expected, RARPH data distributions, both for actions and states, is in-between the ones obtained with random policies and random actions, as shown on figure \ref{hist_repart}.

\subsection{Prediction error analysis}

%%% Redundant Arm %%%

\begin{wraptable}{r}[5mm]{0.7\textwidth}
\scalebox{0.6}{
\begin{tabular}{| c | c | c | c | c|}
    \hline
    \multicolumn{5}{| c |}{Redundant Arm}\\\hline
    Initialization episodes & Prediction horizon & Random policies & Random actions & RARPH\\\hline

    \multicolumn{1}{| c |}{\multirow{3}{*}{5}}
    & 1     & 0.011 $\pm$ 0.029 & 0.002 $\pm$ 0.002 & 0.002 $\pm$ 0.002\\\cline{2-5}
    & 20    & 0.236 $\pm$ 0.673 & 0.047 $\pm$ 0.042 & 0.046 $\pm$ 0.041\\\cline{2-5}
    & 250   & 16.34 $\pm$ 5.855 & 1.084 $\pm$ 0.350 & 1.071 $\pm$ 0.366\\\hline
    
    \multicolumn{1}{| c |}{\multirow{3}{*}{10}}
    & 1     & 0.002 $\pm$ 0.002 & 0.002 $\pm$ 0.002 & 0.002 $\pm$ 0.002\\\cline{2-5}
    & 20    & 0.041 $\pm$ 0.035 & 0.044 $\pm$ 0.037 & 0.044 $\pm$ 0.039\\\cline{2-5}
    & 250   & 1.002 $\pm$ 0.314 & 1.066 $\pm$ 0.342 & 1.051 $\pm$ 0.359\\\hline
    
    \multicolumn{1}{| c |}{\multirow{3}{*}{15}}
    & 1     & 0.002 $\pm$ 0.002 & 0.002 $\pm$ 0.002 & 0.002 $\pm$ 0.002\\\cline{2-5}
    & 20    & 0.038 $\pm$ 0.034 & 0.041 $\pm$ 0.034 & 0.040 $\pm$ 0.035\\\cline{2-5}
    & 250   & 0.917 $\pm$ 0.301 & 1.050 $\pm$ 0.322 & 1.002 $\pm$ 0.328\\\hline
    
    \multicolumn{1}{| c |}{\multirow{3}{*}{20}}
    & 1     & 0.002 $\pm$ 0.002 & 0.002 $\pm$ 0.002 & 0.002 $\pm$ 0.002\\\cline{2-5}
    & 20    & 0.033 $\pm$ 0.032 & 0.040 $\pm$ 0.034 & 0.039 $\pm$ 0.034\\\cline{2-5}
    & 250   & 0.918 $\pm$ 0.287 & 1.051 $\pm$ 0.340 & 1.019 $\pm$ 0.334\\\hline
\end{tabular}}
\captionsetup{justification=centering}
\caption{Mean prediction error on Redundant Arm}
\label{quanti_mean_pred:redundant_arm}
\vspace{-5mm}
\end{wraptable}

%To confirm the results obtained by analyzing directly the data distributions, we measure the model prediction error. Firstly, we analyze the mean prediction error and the associated standard deviation on each environment for various initialization episode budgets and various prediction horizons. Secondly and analogously, we analyze qualitatively the mean prediction error on example trajectories (obtained from NS runs and selected for their challenging behaviours).

%%%%%%%%%%% Mean and std prediction errors and graphs over all reps %%%%%%%%%%%
\par 
The results of this section are obtained by averaging the mean prediction error over 10 repetitions on NS examples trajectories. Model is either used for 1-step predictions or recursively for 20 or $H$ steps, $H$ being the task horizon.

Looking at Table \ref{quanti_mean_pred:ball_in_cup}, we observe that the mean prediction errors and standard deviations are smaller by a factor of around two on all initialization budgets and all prediction horizons between random policies and random actions. Qualitative results are available in appendix A. This validates previous results that seemed to show that random policies are better at gathering initial data for training a dynamics model on the Ball In Cup. Moreover, RARPH initialization performs better than random actions, but does not run up to random policies prediction quality.

%%% Redundant Arm No Walls%%%

\begin{wraptable}{r}[5mm]{0.7\textwidth}
\vspace{-5mm}
\scalebox{0.6}{
\begin{tabular}{| c | c | c | c | c|}
    \hline
    %\multicolumn{2}{| c |}{\multirow{2}{*}{Multi-col-row}} & x & x\\\hline
    \multicolumn{5}{| c |}{Redundant Arm No Walls}\\\hline
    %Initialization episodes & Prediction horizon & Random Policies & Random Actions\\\hline
    Initialization episodes & Prediction horizon & Random policies & Random actions & RARPH\\\hline

    \multicolumn{1}{| c |}{\multirow{3}{*}{5}}
    & 1     & 0.038 $\pm$ 0.286 & 0.004 $\pm$ 0.005 & 0.004 $\pm$ 0.004\\\cline{2-5}
    & 20    & 1.145 $\pm$ 7.778 & 0.082 $\pm$ 0.078 & 0.075 $\pm$ 0.076\\\cline{2-5}
    & 250   & 20.37 $\pm$ 50.97 & 2.558 $\pm$ 0.613 & 2.445 $\pm$ 0.177\\\hline
    
    \multicolumn{1}{| c |}{\multirow{3}{*}{10}}
    & 1     & 0.003 $\pm$ 0.004 & 0.003 $\pm$ 0.004 & 0.003 $\pm$ 0.004\\\cline{2-5}
    & 20    & 0.063 $\pm$ 0.063 & 0.065 $\pm$ 0.069 & 0.067 $\pm$ 0.062\\\cline{2-5}
    & 250   & 2.864 $\pm$ 0.770 & 2.491 $\pm$ 0.153 & 2.521 $\pm$ 0.236\\\hline
    
    \multicolumn{1}{| c |}{\multirow{3}{*}{15}}
    & 1     & 0.003 $\pm$ 0.003 & 0.003 $\pm$ 0.003 & 0.003 $\pm$ 0.004\\\cline{2-5}
    & 20    & 0.059 $\pm$ 0.059 & 0.056 $\pm$ 0.056 & 0.063 $\pm$ 0.068\\\cline{2-5}
    & 250   & 2.534 $\pm$ 0.304 & 2.532 $\pm$ 0.175 & 2.405 $\pm$ 0.144\\\hline
    
    \multicolumn{1}{| c |}{\multirow{3}{*}{20}}
    & 1     & 0.003 $\pm$ 0.003 & 0.003 $\pm$ 0.003 & 0.003 $\pm$ 0.003\\\cline{2-5}
    & 20    & 0.058 $\pm$ 0.059 & 0.057 $\pm$ 0.055 & 0.060 $\pm$ 0.062\\\cline{2-5}
    & 250   & 2.485 $\pm$ 0.291 & 2.476 $\pm$ 0.198 & 2.458 $\pm$ 0.116 \\\hline
\end{tabular}}
\captionsetup{justification=centering}
\caption{Mean prediction error on Redundant Arm without walls}
\label{quanti_mean_pred:redundant_arm_no_walls}
\vspace{-5mm}
\end{wraptable}

\par 
On the Redundant Arm environment, we observe that random actions and RARPH have a better prediction performance (Table \ref{quanti_mean_pred:redundant_arm}), RARPH even slightly outperforming random actions. This is especially true for small episodes budget, where random policies suffer most from the lack of data due to early stopping. Indeed, when removing the walls, we observe that random policies and random actions perform equally better, except for the smallest budget scenario where self-collision must play an important part in the early stopping of random policies.

%%%%%%%%%%%%%%%%%%%%%%%%%%%%%%%%%%%%%%%%%%%%%%%%%%%%%%%%%%%%%%%%%%%%%%%%%%%%%%%%%%%%%%
%%%%%%%%%%%%%%%%%%%%%%%%%%%%%%%%%%%%%%%%%%%%%%%%%%%%%%%%%%%%%%%%%%%%%%%%%%%%%%%%%%%%%%
%%%%%%%%%%%%%%%%%%%%%%%%%%%%%%%%%%% CONCLUSION %%%%%%%%%%%%%%%%%%%%%%%%%%%%%%%%%%%%%%%
%%%%%%%%%%%%%%%%%%%%%%%%%%%%%%%%%%%%%%%%%%%%%%%%%%%%%%%%%%%%%%%%%%%%%%%%%%%%%%%%%%%%%%
%%%%%%%%%%%%%%%%%%%%%%%%%%%%%%%%%%%%%%%%%%%%%%%%%%%%%%%%%%%%%%%%%%%%%%%%%%%%%%%%%%%%%%

\section{Conclusion}

In this paper we have shown the importance of comprehending the task dynamics and of selecting the data gathering initialization method beforehand. We compared two initialization methods employed in the state of the art and compared their performance on three different experimental setups. Results show that various factors like sparse interactions or early-stopping criterion can be detrimental to one method or another, suggesting hybrid approaches. The hybrid approach does not yield best results on all tasks, but its performances are more regular than other techniques.

\newpage

\acks{This work has received funding from the European Union's Horizon 2020 research and innovation program under grant agreement no 869855 (Project 'SoftManBot').}

\appendix
\section*{Appendix A.}
\label{app:quali_mean_pred}

\begin{figure}[h!]
    \begin{subfigure}{.329\textwidth}
        \centering
        \includegraphics[scale=0.3]{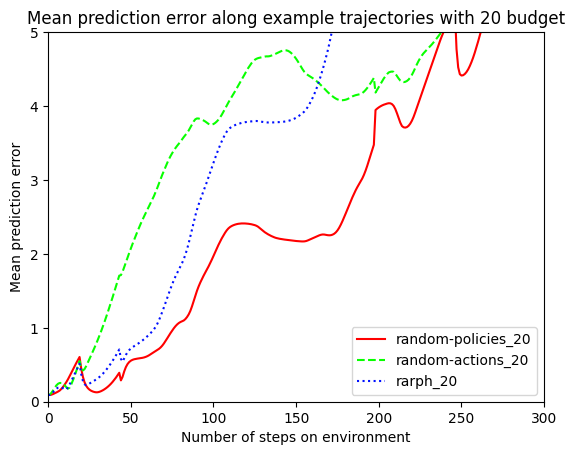}
        \caption{Ball In Cup}
        \label{quali_mean_pred:bic}
    \end{subfigure}
    \begin{subfigure}{.329\textwidth}
        \centering
        \includegraphics[scale=0.3]{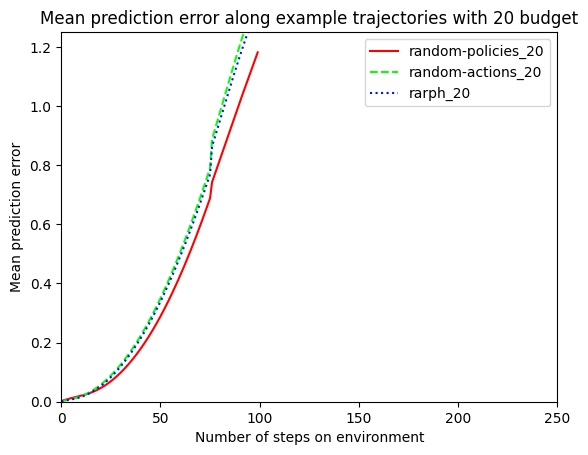}
        \caption{Redundant Arm}
        \label{quali_mean_pred:ra}
    \end{subfigure}
    \begin{subfigure}{.329\textwidth}
        \centering
        \includegraphics[scale=0.3]{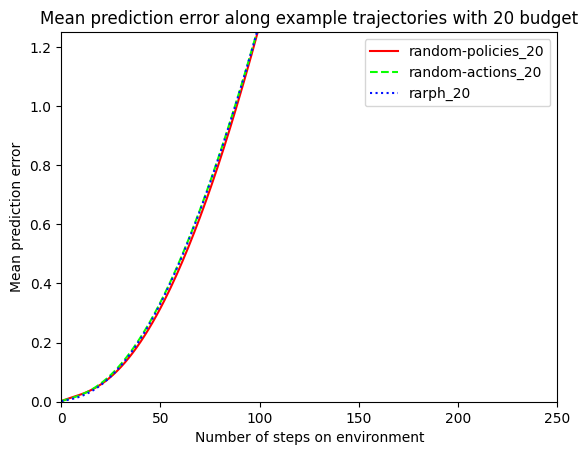}
        \caption{RA no walls}
        \label{quali_mean_pred:ranw}
    \end{subfigure}
    \captionsetup{justification=centering}
    \caption{Initialization methods mean prediction error over whole state-space on considered environments on NS archive trajectories on a 20 episode budget}
    \label{quali_mean_pred}
\end{figure}

%\newpage

%\appendix
%\section*{Appendix A.}

% Note: in this sample, the section number is hard-coded in. Following
% proper LaTeX conventions, it should properly be coded as a reference:

%In this appendix we prove the following theorem from
%Section~\ref{sec:textree-generalization}:

\vskip 0.2in
\bibliography{sample}

\end{document}